\documentclass[letterpaper, 10 pt, conference]{ieeeconf}  
\IEEEoverridecommandlockouts                              

\usepackage{graphics} 
\usepackage{graphicx}
\usepackage{placeins}
\usepackage{epsfig} 
\usepackage{amsmath} 
\usepackage{amssymb}  
\usepackage{booktabs}
\usepackage{array}
\newcommand\hidden[1]{} 
\usepackage{mathtools}
\newtagform{hidden}[\hidden]{}{}

\usepackage{xcolor}
\definecolor{myteal}{HTML}{008080}
\definecolor{mygrey}{HTML}{a9a9a9}

\usepackage{algorithm}
\usepackage[noend]{algpseudocode} 
\algnewcommand\AAND{\textbf{ and }}
\algnewcommand\Or{\textbf{ or }}

\usepackage{citesort}
\usepackage{url}
\usepackage{breakurl}
\usepackage[breaklinks]{hyperref}
\usepackage{caption}
  \usepackage{censor}                                                                                                                                                                  
\usepackage{titlesec}

\titlespacing*{\section}
{0pt}{1.2ex plus .5ex minus .2ex}{0.8ex plus .2ex}

\titlespacing*{\subsection}
{0pt}{0.85ex plus .3ex minus .2ex}{0.6ex plus .2ex}

\titlespacing*{\subsubsection}
{0pt}{0.8ex plus .2ex}{0.5ex}

\titlespacing*{\paragraph}
{0pt}{0.8ex}{0.5em}

\DeclareMathAlphabet{\pazocal}{OMS}{zplm}{m}{n}

\usepackage{xspace}
\newcommand{\videoexptimestampone}{1:28\xspace}
\newcommand{\videoexptimestamptwo}{2:11\xspace}
\newcommand{\videoexptimestampthree}{2:56\xspace}

\newcommand{\percentageTaskAOptimizedBetterThanPlanar}{21\%\xspace}
\newcommand{\percentageTaskAOptimizedBetterThanFranchi}{7\%\xspace}
\newcommand{\percentageTaskBOptimizedBetterThanPlanar}{28\%\xspace}
\newcommand{\percentageTaskBOptimizedBetterThanFranchi}{17\%\xspace}
\newcommand{\percentageTaskBOptimizedBetterThanPlanarQuad}{5\%\xspace}

\newcommand{\percentageCmaesAfterBoTaskA}{2.9\%\xspace}
\newcommand{\percentageCmaesAfterBoTaskB}{3.3\%\xspace}

\newcommand{\evalsBoTaskA}{386\xspace}
\newcommand{\evalsCMATaskA}{224\xspace}
\newcommand{\totalevalsTaskA}{610\xspace}

\newcommand{\evalsBoTaskB}{647\xspace}
\newcommand{\evalsCMATaskB}{191\xspace}
\newcommand{\totalevalsTaskB}{838\xspace}

\def \*#1 {\mathbf{#1}}
\def \@#1 {\mathbb{#1}}

\DeclareMathAlphabet{\mathpzc}{OT1}{pzc}{m}{it}
\usepackage[font=footnotesize]{caption}
\usepackage[font=footnotesize]{subcaption}

\newcolumntype{C}[1]{>{\centering\arraybackslash}p{#1}}
\newcolumntype{M}[1]{>{\raggedright\arraybackslash}p{#1}}

\newcolumntype{L}[1]{>{\raggedright\let\newline\\\arraybackslash\hspace{0pt}}m{#1}}
\newcolumntype{S}[1]{>{\centering\let\newline\\\arraybackslash\hspace{0pt}}m{#1}}
\newcolumntype{R}[1]{>{\raggedleft\let\newline\\\arraybackslash\hspace{0pt}}m{#1}}

\usepackage[nolist,nohyperlinks]{acronym}

\makeatletter
\def\endthebibliography{%
	\def\@noitemerr{\@latex@warning{Empty `thebibliography' environment}}%
	\endlist
}
\makeatother

\acrodef{mav}[MAV]{Micro Aerial Vehicle}
\acrodef{fov}[FoV]{Field of View}
\acrodef{tsp}[TSP]{Traveling Salesman Problem}
\acrodef{rps}[RPS]{Rotations Per Second}
\acrodef{rl}[RL]{Reinforcement Learning}
\acrodef{cmaes}[CMA-ES]{Covariance Matrix Adaptation Evolution Strategy}
\acrodef{bo}[BO]{Bayesian Optimization}
\acrodef{ppo}[PPO]{Proximal Policy Optimization}
\acrodef{mse}[RMSE]{Root Mean Squared Error}

\title{\LARGE \bf
Performance-guided Task-specific Optimization for Multirotor Design
}

\author{
Etor Arza, Welf Rehberg, Philipp Weiss, Mihir Kulkarni and Kostas Alexis
\thanks{This research was supported by the European Commission through the European Union’s Horizon Europe Research and Innovation Programme, under the Grant Agreement No. 101119774 SPEAR.}
\thanks{The authors are with the Autonomous Robots Lab, Norwegian University of Science and Technology (NTNU), O. S. Bragstads Plass 2D, 7034, Trondheim, Norway {\tt\small etor.arza@ntnu.no}}
}

\begin{document}

\maketitle
\thispagestyle{empty}
\pagestyle{empty}

\begin{abstract}
This paper introduces a methodology for task-specific design optimization of multirotor Micro Aerial Vehicles. 
By leveraging reinforcement learning, Bayesian optimization, and covariance matrix adaptation evolution strategy, we optimize aerial robot designs guided exclusively by their closed-loop performance in a considered task. Our approach systematically explores the design space of motor pose configurations while ensuring manufacturability constraints and minimal aerodynamic interference. Results demonstrate that optimized designs achieve superior performance compared to conventional multirotor configurations in agile waypoint navigation tasks, including against fully actuated designs from the literature. We build and test one of the optimized designs in the real world to validate the sim2real transferability of our approach. 
\end{abstract}

\section{Introduction}

Robot performance depends on both the mechatronic form (design) and its sensorimotor policy (controller).
Conventionally, robots are first designed considering metrics directly associated with their mechatronic configuration and engineering intuition. For \acp{mav}, common design metrics include thrust-to-weight ratio, moments of inertia, power consumption at hover, and more. While informative, these metrics do not directly command how the system will perform in a certain class of navigation missions. 
Some aerial vehicles have been designed to perform better specifically for a given task, such as aerial robots for object manipulation with an increased number of degrees of freedom and thrust~\cite{olleroPresentFutureAerial2022} or a fully actuated hexarotor with tilted propellers with increased stability~\cite{franchiFullposeTrackingControl2018}.
Still, these systems were designed conventionally, missing the opportunity to holistically co-optimize both the physical design and the control policy in one unified computational process; something that can potentially improve performance with a design tuned for a specific task, possibly beyond what is possible by human intuition.

\begin{figure}
	\centering
	\includegraphics[width=0.9\columnwidth]{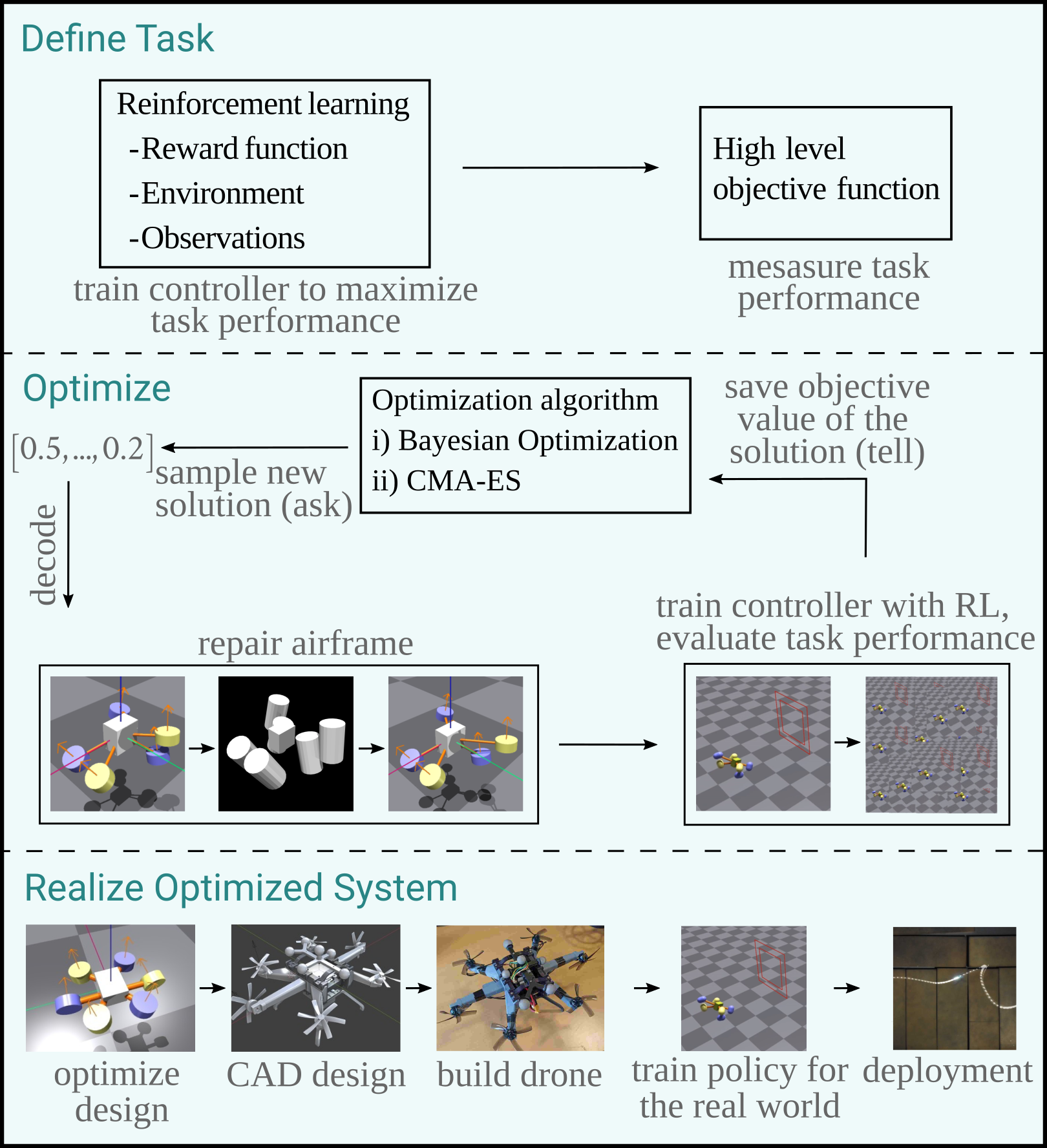}
	\caption{Summary diagram of the full optimization pipeline.}
	\label{fig:diagram_title}
    \vspace{-2ex}
\end{figure}

This work offers a computational pathway towards task-specific \ac{mav} co-optimization by incorporating iterative simulation and performance evaluation. 
Our method simulates \ac{mav} designs and measures their learned closed-loop performance in specific navigation tasks, thus iteratively improving the design based on task performance.
This paper focuses on optimizing hexarotors as they offer a sufficiently large optimization space to demonstrate the merit of our method (with a number of motors that enables full actuation), but we also test the methodology on quadrotors.
Specifically, we optimize the pose of the motors on a hexarotor \ac{mav} to maximize the navigation performance of the system for a specific waypoint navigation task (see Figure~\ref{fig:diagram_title}). 
Importantly, by navigation task we do not refer to any particular maneuver or fixed waypoints but to a class of waypoint missions sampled with specific structure and statistical properties. 

To achieve this, we first define the search space, which consists of a set of feasible hexarotor designs that satisfy certain fabrication constraints including avoiding propeller self-collisions. 
Then, an end-to-end \ac{rl} procedure is developed that trains the controller for a candidate \ac{mav} design while considering its nonlinear dynamics envelope.
Once trained, the performance of each design is evaluated on a target waypoint navigation task, where the waypoints are randomly sampled from a specific distribution.
Based on the results of this evaluation, we sample a new design with \ac{bo}~\cite{frazierTutorialBayesianOptimization2018} that is then trained and tested again.
By iterating this cycle, we find increasingly better-performing novel hexarotor designs that are specifically optimized for the considered waypoint navigation task.
Finally, we refine the airframe design obtained from the previous process using the \ac{cmaes} algorithm.

Results show that optimized designs robustly outperform the planar hexarotor and other tilted hexarotors in the literature for the task that they are optimized for. To further validate our results, we built and tested one of the optimized designs in the real world, confirming the benefits of computational co-optimization of the airframe and policy, and demonstrating that performance is close to that of the simulation. 
A video presenting simulation and experimental results is available at \url{https://youtu.be/V6w_DTKWvtc}.
The method shall be open-sourced upon acceptance.

The remainder of the paper is organized as follows. Section~\ref{sec:related_work} reviews related work in performance-guided co-optimization and \ac{mav} design optimization. Section~\ref{sec:methodology} presents the methodology for task-specific design optimization. Section~\ref{sec:experimentation} describes experimental results, where we also compare optimized designs against existing designs in the literature. Section~\ref{sec:sim2Real} addresses sim2real transfer through physical implementation. Section~\ref{sec:conclusion} draws conclusions.

\section{Related work}
\label{sec:related_work}

\subsection{Performance-guided co-optimization of design and control of robots}

Most works on performance-guided co-optimization of design and control remain limited to simulation, including studies on legged robots~\cite{10.1145/3414685.3417831,a.belmonte-baezaMetaReinforcementLearning2022}, soft robots~\cite{bhatia2021evolution,liReinforcementLearningFreeform2024}, and object manipulators~\cite{toussaintCooptimizingRobotEnvironment2021}.
A limited set of approaches have progressed to real-world implementation, demonstrated among others in both legged robots~\cite{liaoDataefficientLearningMorphology2019,lanTimeEfficiencyOptimization2022} and soft robots~\cite{6096440,strgarEvolutionLearningDifferentiable2024}.


Mannam et al.~\cite{mannamDesignControlCooptimization2024} introduced a procedure to optimize the shape of soft robotic hands for object manipulation with policy learning and genetic algorithms.
Their approach involves a policy transfer procedure where a previously optimized policy is fine-tuned for the new design: the original design is transformed iteratively into the new one with smaller transformations at a time and policy retraining in between each transformation step. 
They show good sim2real transfer, where optimized designs outperform the human-made ones for grasping tasks in the real world.

Ha et al.~\cite{haJointOptimizationRobot2017} propose a framework for jointly optimizing robot design and control using the implicit function theorem, relating design parameters (e.g., link lengths, actuator placements) to motion parameters (e.g., joint trajectories and contact forces). They validate their method in simulation on a manipulator and a quadruped, and demonstrate sim2real transfer on the quadruped, where the required torques were observed to be lower than for the baseline. However, the approach is limited to optimizing for a specific trajectory.


Belmonte et al.~\cite{a.belmonte-baezaMetaReinforcementLearning2022} consider a co-optimization procedure to optimize quadruped robot leg designs by searching over link lengths and gear ratios with \ac{cmaes}. 
They train an \ac{rl} policy that generalizes across different robot designs and can be further fine-tuned for each specific morphology.
Designs are evaluated on velocity tracking accuracy, joint torque minimization, and mechanical power consumption.
While their framework demonstrates that optimized designs outperform baselines, results remain limited to simulation environments without validation on physical hardware.

\subsection{\ac{mav} design optimization}


Ang et al.~\cite{angMultiobjectiveEvolutionDrone2025} propose a method to optimize \acp{mav} in a multi-objective setting, where the goal is to maximize thrust-to-weight ratio and maneuverability, while minimizing size. 
However, as the optimization is guided by these generic metrics, it does not necessarily produce designs tailored to the requirements of a specific navigation task.

Bosio et al.~\cite{bosioAutomatedLayoutControl2025} propose a method to optimize the multirotor layout of multi-Unmanned Aerial Vehicle (UAV) systems (specifically several \acp{mav} acting together) for robustness to disturbances for a specific payload.
The approach is validated in hardware experiments. However, the optimization is limited to a 2D placement problem of standard quadrotors.
Additionally, the controller only assumes a linearized dynamics model. 


Du et al.~\cite{duComputationalMulticopterDesign2016} present a design and control optimization framework for multirotor \acp{mav} that, given an initial configuration and desired LQR-controller weight matrices $\mathbf{Q}$ and $\mathbf{R}$, returns optimized design parameters and a controller gain matrix $\mathbf{K}$ (by finding equilibrium points to linearize around).
They demonstrate successful sim2real transfer with optimized multirotor designs capable of handling larger payloads than baseline planar quadrotors.
However, it does not exploit the full dynamic potential of the configurations since the controller assumes an LQR structure.
Moreover, the user-defined objectives are only calculated based on the system's steady state and are restricted to bi-convex functions. 

Zhao et al.~\cite{zhaoAutomaticCodesignAerial2022} propose a similar approach for fixed-wing UAVs, which can co-optimize wing structure and motor-propeller placement. However, their approach (a) also relies on LQR control thus fails to capture the full nonlinear dynamics of the UAVs, while (b) remains limited to simulation. 


Muff et al.~\cite{muffUnconventionalHexacoptersEvolution2025} propose a framework for optimizing hexarotors based on performance alone, showing that optimized designs outperform the baseline planar hexarotor in tracking specific trajectories. However, the study only considers fixed waypoints, is limited to comparisons with only the planar hexarotor and is conducted solely in simulation.

Departing from limitations of prior approaches, this work optimizes \acp{mav} driven by their performance in navigation tasks (defined as particular waypoint distributions) rather than a single trajectory or simple body metrics such as thrust-to-weight ratio, while leveraging the full nonlinear dynamics of the system through \ac{rl}.
This enables the discovery of unconventional hexarotor configurations that outperform not only standard planar designs but also fully actuated designs from the literature, demonstrating the value of task-specific co-optimization guided by closed-loop performance. 

\section{Task-specific Design Optimization}
\label{sec:methodology}

The proposed methodology for performance-driven task-specific multirotor airframe design optimization is presented.

\subsection{Encoding a design}
\label{sec:encoding_designs}

Each hexarotor design is represented as a real-valued vector $\xi \in [0,1]^{15} \subset \mathbb{R}^{15}$, where the parameters represent the poses of three of the motors in the airframe assuming a symmetry to the $xz$-plane in the body frame.
Without loss of generality, in this work, we consider a single motor-propeller model combination as a simplification. 
In the following, we define how the parameter vector $\xi$ is decoded into a valid airframe design.
The decoding is done in 2 steps.

\begin{figure}[t]
	\centering
	\includegraphics[width=0.70\linewidth]{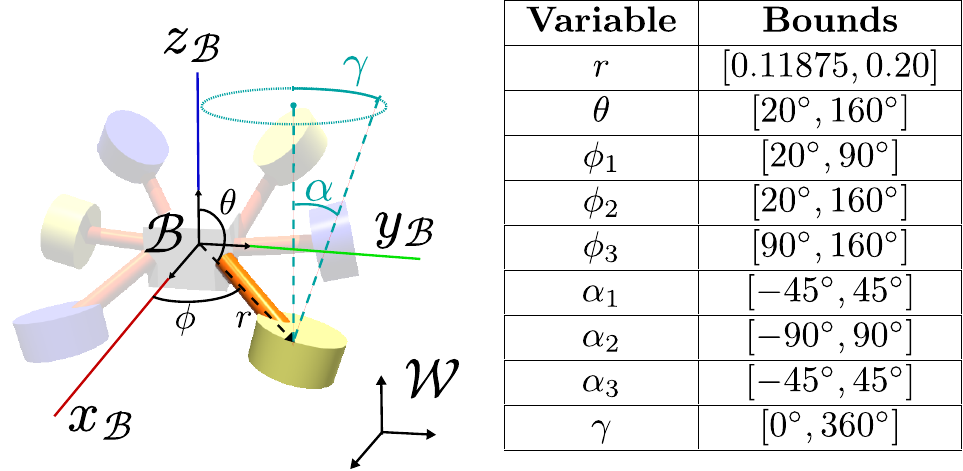}
	\caption{Position and orientation of each motor (with $xz$ plane symmetry). Translation is given in polar coordinates $(r,\theta,\phi)$ and orientation is defined with two Euler angles $(\alpha,\gamma)$. The table shows the bounds of the variables.}
	\label{fig:spherical-coordinates}

\end{figure} 

\subsubsection*{Step 1: Decode into airframe parameters}
The encoding contains three tuples $(r,\theta,\phi,\alpha,\gamma)$ of size 5, where each tuple represents the position and orientation of a single motor, while the other three motors are symmetric to the $xz$\nobreakdash-plane in the body frame.
The first three variables $(r,\theta,\phi)$ represent the position of a motor in spherical coordinates, and the next two $(\alpha,\gamma)$ represent the orientation with two Euler angles (see Figure~\ref{fig:spherical-coordinates} for a diagram).
The spinning direction of the propellers is alternated between clockwise and counter-clockwise for each consecutive motor.
We opted for designs with a symmetry on the $xz$-plane.
This makes the optimization more tractable by halving the number of optimizable parameters and removing some unreasonable design such as designs with no motors on one side of the $xz$-plane.

Not every parameter combination produces a design that can hover.
The table in Figure~\ref{fig:spherical-coordinates} shows the maximum and minimum values of the variables.
To reduce the number of redundant and bad designs, we limit $\phi$ and $\alpha$ to different subranges for different motors, as shown in the table.
While these bounds could be relaxed in favor of relying solely on the hovering constraint, this would substantially increase the proportion of infeasible designs evaluated during optimization.

\subsubsection*{Step 2: Repair Airframe}

Not all airframe configurations are feasible in practice, so we introduce several constraints to ensure physical realizability and reduce the unmodeled aerodynamic interference in the process.
We consider the following:
a)~two propellers must not collide with each other,
b)~no propeller should intersect with the electronics cage (i.e., central airframe part containing electronics and battery), c)~propeller airflows must not interfere significantly with each other or with the electronics cage, as such interferences reduce efficiency and introduce aerodynamic effects not captured by the simulation (for negligible effects vertically placed propellers should be at least at 2 diameters of distance from each other~\cite{FAA_AIM_Ch7_Sec4_2025}), and
d)~symmetry must be maintained with respect to the $xz$\nobreakdash-plane, where we place three motors and propellers on each side.

\begin{figure}
	 \captionsetup[subfigure]{labelformat=parens, labelsep=space}
	 \renewcommand{\thesubfigure}{\arabic{subfigure}}
	\centering
	\begin{subfigure}[b]{0.11\textwidth}
		\includegraphics[width=\textwidth]{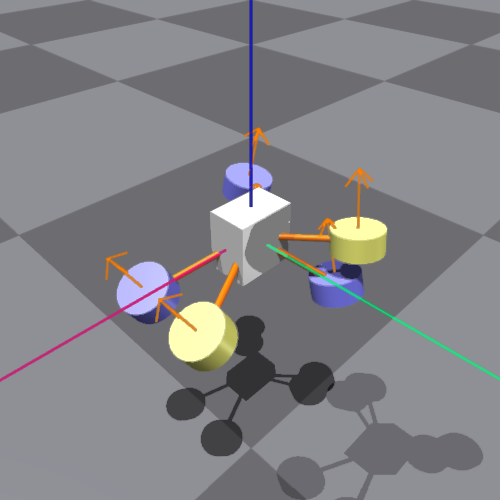}
		\caption{ }
		\label{fig:repair_1}
	\end{subfigure}
	\hfill
	\begin{subfigure}[b]{0.11\textwidth}
		\includegraphics[width=\textwidth]{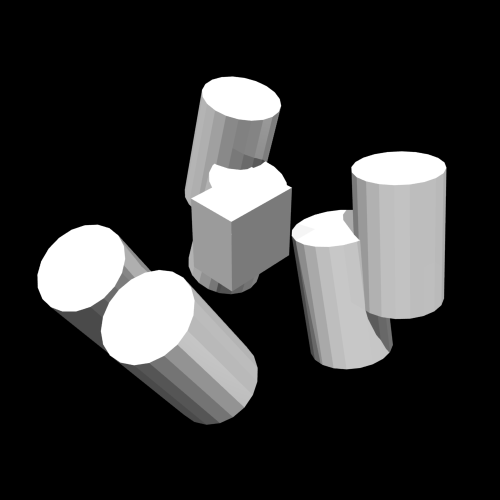}
		\caption{ }
		\label{fig:repair_2}
	\end{subfigure}
	\hfill
	\begin{subfigure}[b]{0.11\textwidth}
		\includegraphics[width=\textwidth]{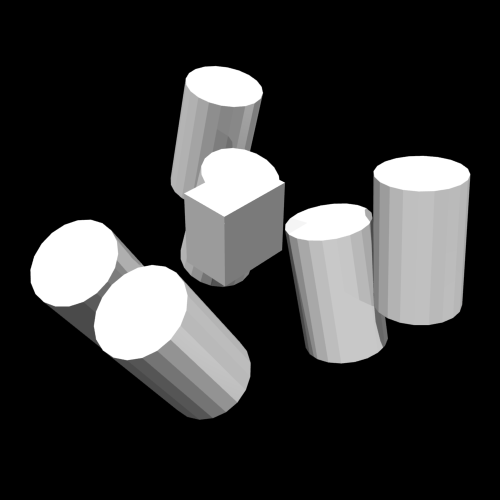}
		\caption{ }
		\label{fig:repair_3}
	\end{subfigure}
	\hfill
	\begin{subfigure}[b]{0.11\textwidth}
		\includegraphics[width=\textwidth]{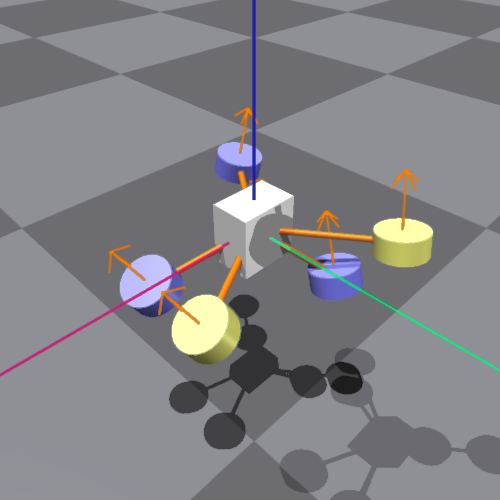}
		\caption{ }
		\label{fig:repair_4}
	\end{subfigure}
	\caption{Steps of the repair procedure to ensure a feasible airframe design. (1)~Initial design: interfering airflows. (2)~Retrieve the pose of the motors and add the margins representing the airflow. (3)~Solve \textit{Optimization Problem: repair airframe design} to find the minimum translation for each motor that separates them from each other and from the electronics cage. (4)~Extract the new positions of the motors.}
	\label{fig:repair_procedure}
    \vspace{-2ex}
\end{figure}

To ensure that every airframe satisfies these four constraints, we consider a repair procedure that translates the motors until the design is feasible (see Figure~\ref{fig:repair_procedure}).
Note that since we only translate the motors and do not consider changes in orientation during repair, the linear forces produced by the system remain unchanged, and only the torques are affected by the repair procedure.
In practice, we model it as an object collision problem, where the goal is to minimize the translation applied to the motors subject to the objects not colliding with each other.
This object collision problem models all of the previous four constraints (including no airflow interference).


We model the electronics cage as a $9 \times 6 \times 9$~cm cuboid, the propeller-motor combination as a disk with diameter equal to the propeller diameter plus 2~mm of margin, and the propeller airflow as a cylinder with height four times the propeller radius (equivalent to a separation of two diameters between vertically stacked propellers)~\cite{FAA_AIM_Ch7_Sec4_2025}.
A design satisfies constraints a), b) and c) when these objects are not colliding with each other.
More specifically, the repair procedure is carried out via the following Optimization Problem, where the goal is to find the most similar airframe (in terms of minimum translation of the motors) while keeping symmetry and ensuring no object collisions.

We use penetration depth~\cite{panFCLGeneralPurpose2012} as a continuous measure of collision.
The penetration depth between two objects is defined as the minimum translation distance needed to separate them~\cite{panFCLGeneralPurpose2012}.
When the penetration depth is 0, the objects are not colliding.
We use the FCL library~\cite{panFCLGeneralPurpose2012} to compute the penetration depth and the differential evolution optimizer in SciPy~\cite{virtanenetal.SciPy10Fundamental2020}.

 \begin{small}
 	\begin{equation*}
 		\label{prb:repair}
 		\begin{aligned}
 			\hline
 			& \textbf{Optimization Problem: Repair airframe design} \\
 			\hline
 			& \underset{\Delta T_{i}}{\text{Minimize}} \quad \sum_{i=1}^{6} \|\Delta T_{i}\|_2 \\
 			& \text{Subject to} \\
 			&  (C_i + \Delta T_{i}) \cap (C_j + \Delta T_{j}) = \emptyset ,\\
 			& (C_i + \Delta T_{i})  \cap \text{electronics\_cage} = \emptyset ,\\
 			& \forall \, i, j \in \{1, \ldots, 6\}, \, i \neq j; \\
 			& (x_i, y_i, z_i) = (x_{i+3}, -y_{i+3}, z_{i+3}), \quad \forall i \in \{1,2,3\}; \\
 			& \text{where } C_i \text{ is a cylinder centered at and oriented as motor}\\
 			& i \text{ with length } 4r_p \text{ and radius } r_p + 10^{-3} \text{ (}r_p \text{ is the radius of}\\
 			& \text{the propeller) and } \Delta T_{i} = (x_{i}, y_{i}, z_{i}) \text{ is the translation} \\ 
 			& \text{applied to motor } i. \\
 			\hline 
 		\end{aligned}
 	\end{equation*}
 	\vspace{-2ex}
 \end{small}

\subsection{Simulation}
\label{sec:simulation}

We simulate the airframe designs using the Aerial Gym Simulator~\cite{aerial_gym_github}, a high-throughput GPU parallel simulator for \acp{mav} based on NVIDIA Isaac Gym~\cite{makoviychukIsaacGymHigh2021}.
An airframe design that performs well in simulation does not necessarily perform well in the real world.
We took the following steps to mitigate this issue.
As mentioned in the previous section, we only consider airframe designs with non-interfering airflows, which are not modeled by Aerial Gym (and broadly represent a complicated process that is not integrated into most established robotics simulators).
We identified the time constant of the motor-propeller combination (see Section~\ref{sec:sim2Real}) and simulate them in a first-order model, setting the minimum and maximum \ac{rps} to 83 and 400, as we experimentally observed the first-order approximation to be fairly accurate between those \ac{rps} values.

We estimate the mass and inertia of each airframe design that we evaluate.
We calculate the optimal arm attachment placements such that the arm length is as low as possible. 
CAD was used to estimate the inertia and mass of the electronics cage/battery, and the arms (which depend on arm length).
Motors and propellers are considered point masses for inertia calculation. 


\subsection{Task definition}
\label{sec:task_definition}
The two tasks considered in this paper are waypoint navigation tasks motivated by two specific scenarios.
By placing waypoints along a mostly straight path with small lateral deviations, we can define a task that requires continuous smooth dodging of obstacles in forward flight.
This reflects the maneuvering required when flying inside a forest under the canopy.
Alternatively, by introducing rare but large lateral offsets in the waypoint placement, we can define a task that demands last-resort emergency-avoidance maneuvers.
These two scenarios are formalized as \texttt{task A} and \texttt{task B} below.

\begin{figure}
	\centering
	\noindent
	\begin{minipage}{0.35\linewidth}
		\centering
		\captionof{table}{Ranges for the random initial state.}
		\label{tab:initial_state_ranges}
		\scriptsize
		\begin{tabular}{|@{}c@{}|@{}c@{}|}
			\hline
			\textbf{Variable} & \textbf{Range} \\ \hline
			$x_{\mathcal{W}}$ & $[-1.0, -0.5]$~m \\ \hline
			$y_{\mathcal{W}}, z_{\mathcal{W}}$ & $[-0.5, 0.5]$~m \\ \hline
			~roll, pitch~ & $0~^\circ$ \\ \hline
			yaw & $[-30, 30]~^\circ$ \\ \hline
			$v_i$ & $[-0.2, 0.2] ~\textrm{m/s}$ \\ \hline
			$\omega_i$ & ~$[-11.46, 11.46] ~^\circ\textrm{/s}$~ \\ \hline
		\end{tabular}
	\end{minipage}
	\hfill
	\begin{minipage}{0.55\linewidth}
		\centering
		\includegraphics[width=1.0\linewidth]{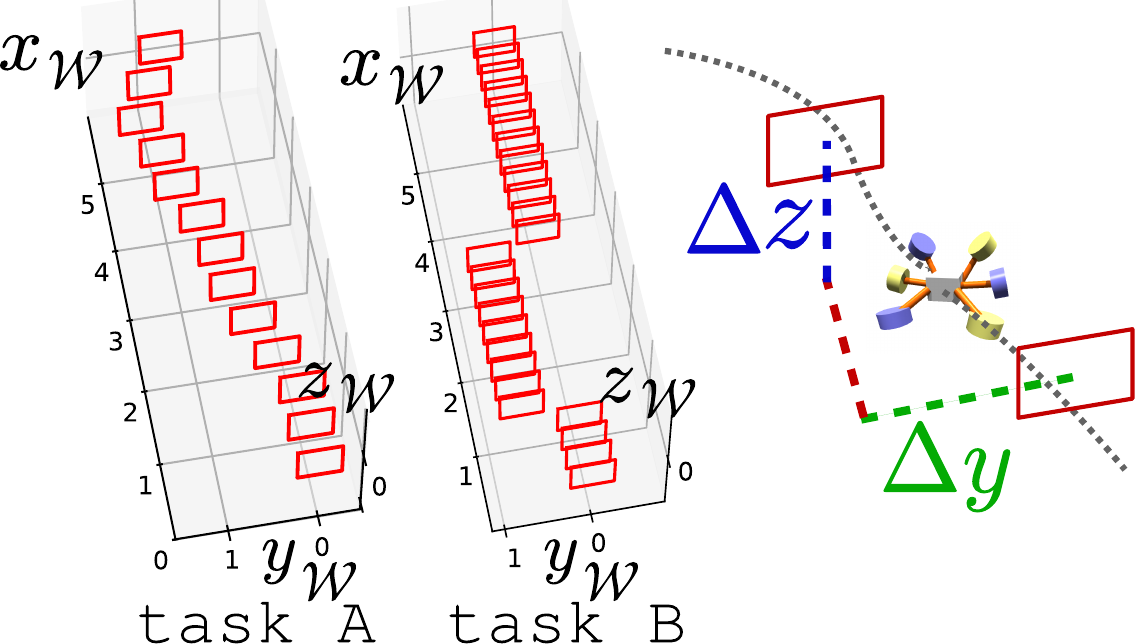}
		\captionof{figure}{
			Two samples of the waypoint probability distributions of \texttt{task A} and \texttt{task B}.
		}
		\label{fig:gatestaskexpltopview}
	\end{minipage}
	\vspace{-3ex}
\end{figure}

In both tasks, the airframe must cross as many square $0.5\times0.5$~m gates (for simplicity called ``waypoints'') as possible in a limited amount of time of 5 seconds, and the waypoints are sampled from a task-specific probability distribution.
We consider that the airframe crossed the waypoint when its center passes through the waypoint, while the waypoint is considered missed when the airframe crosses the plane defining the waypoint outside of the $0.5\times0.5$~m square.
The airframe is initialized at a random position and heading, as described in Table~\ref{tab:initial_state_ranges}, and a new waypoint is sampled each time the airframe crosses or misses the previous waypoint.

\noindent\emph{\texttt{task A}} \emph{forward flight with smooth steering}. Waypoints are sampled at $(0.5, \Delta y, 0)$ meters from the previous waypoint, where ${\Delta y\sim\mathcal{U}([-0.25,0.25])}$, and $\mathcal{U}$ is the uniform distribution, as shown in the left side of Figure~\ref{fig:gatestaskexpltopview}.

\noindent\emph{\texttt{task B}} \emph{forward flight with sudden avoidance}. Waypoints are sampled at $(0.25, \Delta y, \Delta z)$ from the previous waypoint, with 
\vspace{-1.5ex}
\[
(\Delta y, \Delta z) = 
\begin{cases}
	(0, 0) & \text{with probability } 1 - \mathrm{Pr}, \\
	(y_u, z_u) & \text{with probability } \mathrm{Pr},
\end{cases}
\vspace{-0.5ex}
\]
with $y_u \sim \mathcal{U}([-0.7, -0.5] \cup [0.5, 0.7])$, $z_u \sim \mathcal{U}([-0.1, 0.1])$ and $\mathrm{Pr} = 0.05$. 
See the right side of Figure~\ref{fig:gatestaskexpltopview} for a specific sample of \texttt{task~B}.

\subsection{Evaluating task performance}
\label{sec:evaluating_designs}

Task performance is a measure of how well an airframe performs in a waypoint navigation task.
We define \textit{task performance} as the average number of waypoints that an airframe can go through in 5 seconds, minus $10 \times$ the number of waypoints it misses, averaged over $4 \cdot 10^4$ episodes with waypoints sampled with different seeds.

\subsection{Training the controller}
\label{sec:control}


We use Proximal Policy Optimization~\cite{schulman2017proximal} to train the control of arbitrary airframes.
The airframe is controlled with a feed forward neural network with three 64 neuron hidden layers.
The network policy outputs motor commands, given the full state of the airframe as observation.
Specifically, the observation vector $O \in \mathbb{R}^{15}$ consists of:
$O = (p, R_{6D}, v, \omega)$
where $p \in \mathbb{R}^3$ is the relative position vector from the robot to the target waypoint in the world frame, $R_{6D} \in \mathbb{R}^6$ is the 6D continuous representation of the robot's orientation following Zhou et al.~\cite{zhouContinuityRotationRepresentations2019}, $v \in \mathbb{R}^3$ is the robot's linear velocity in the world frame, and $\omega \in \mathbb{R}^3$ is the robot's angular velocity in the body frame.


The reward function is a sum of different terms as shown in the following equation:

\vspace{-3.0ex}
{\footnotesize
\begin{align*}
\mathcal{R} & = \mathcal{R}_{\text{pos}} &\text{ \#  position reward}\\
& - 2 \cdot \max(|\mathcal{R}_{\text{pos}}| \cdot P_{\text{path}}, 0)(1 - c) &\text{ \#  path deviation penalty}\\
& - 2 \cdot \max ((||p_t - s||_2 - d_{\text{best}}) - 0.5, 0) &\text{ \#  no improvement penalty}\\
& - 2 \cdot \max (\|\omega\| - 10, 0) &\text{\# angvel penalty}\\
& - 2 \cdot \max (\alpha_{\text{z}} - 0.5\pi, 0) - 0.01 \cdot \alpha_{\text{y}} &\text{\# orientation penalty}\\
& - 100 \cdot (C_{\text{dist}} + C_{\text{orient}}) \cdot c &\text{ \#  crash penalty}\\
& + 10 \cdot \textit{instantaneous task performance} 
\end{align*}
}
\vspace{-4.5ex}

\noindent
where $p_t$ and $p_{t-1}$ are the positions of the robot in the current and previous time steps, $s$ and $s_{prev}$ are the positions of the next and previous waypoints, $c$ is the curriculum progress increasing from 0 to 1, ${\mathcal{R}_{\text{pos}}} = {(||p_{t-1} - s||_2 - ||p_t - s||_2)(1 - c)}$ is the position reward for moving towards the next waypoint,  ${d_{\text{best}}} = {\min (||p_i - s||_2 \ \ | \ \ i=1,...,t)}$ is the minimum distance from the drone to the next waypoint so far,  $\alpha_{\text{y}},\alpha_{\text{z}}$ measure the angle between the robot's and the world's $y$ and $z$-axes respectively, $C_{\text{dist}}$, $C_{\text{orient}}$ are crash indicator variables, \textit{instantaneous task performance} is ${- 10 \cdot G_{\text{missed}}}  + {1 \cdot G_{\text{crossed}}}$ with $G_{\text{missed}}$ and $G_{\text{crossed}}$ are gate-missed and gate-crossed indicators, and $P_{\text{path}}$ is defined as

\vspace{-2.5ex}
\[
\footnotesize
P_{\text{path}} =
\begin{cases}
0 & \text{if } p_t \in \mathcal{E}_{in}(s_{\text{prev}}, s) \\
1 & \text{if } p_t \notin \mathcal{E}_{out}(s_{\text{prev}}, s) \\
\min(\max(2\dfrac{\max(d_y, d_z)}{(s - p_t) \cdot \hat{\imath}},\, 0), 1) & \text{otherwise}
\end{cases}
\]
\vspace{-1ex}

\noindent
where $\mathcal{E}_{in}(s_{\text{prev}}, s)$ is the corridor connecting two gates, $\mathcal{E}_{out}(s_{\text{prev}}, s)$ is the axis-aligned cuboid containing both gates, $d_y, d_z$ are excess distances outside the allowed corridor and \( \hat{\imath} \) is the unit vector along the x-axis in the world frame.

Out of all these reward components, the instantaneous task performance is the most important: if we sum all of the \textit{instantaneous task performance} components in an episode, averaged over $4 \cdot 10^4$ episodes, we get the \textit{task performance} defined in Section~\ref{sec:evaluating_designs}.
It also has the highest contribution in the reward function in both tasks.\footnote{The contribution of the \textit{instantaneous task performance} is higher than 98\% in both tasks for the planar hexarotor. 
The reward function considered in this paper is a sum of different reward terms. We define the contribution of each reward as the percentage of each reward component in absolute value with respect to the sum of all of the reward terms, averaged over $4\cdot10^4$ episodes.}
The rest of the reward components ensure safe and stable flight and make the learning process more robust.




\subsubsection{Sequential halving}
\label{sec:sequential_halving}
Due to the stochastic nature of reinforcement learning, each training run of the same single airframe design will produce a different policy with a different task performance.
This makes evaluating the task performance of an airframe more challenging: it is not possible to distinguish between an unlucky seed and a bad-performing airframe.
A simplistic approach to address this is to repeatedly train the same design and choose the best-performing policy.
For a more robust assessment, we use sequential halving~\cite{karninAlmostOptimalExploration2013}: we start with 8 runs trained for 800 epochs, continue training the top 6 for another 800, then train the top 4 for 3200 epochs. 
In the end, we keep only the best policy. 
This approach reduces variance while using less compute than fully training all 8 runs.
Even with sequential halving, it is a computationally expensive process and takes approximately 3 hours on an NVIDIA A100 GPU.


%

\subsection{Design optimization}
\label{sec:optimization_explained}

The optimization process seeks to find the best-performing airframe: the airframe that, when trained with the \ac{rl} algorithm, and evaluated on the task, obtains the best possible task performance.
This optimization problem is a black-box stochastic optimization problem, where our ability to evaluate the objective function is limited to the simulation of the airframe design and the evaluation of the task performance, which is a noisy measure (training and evaluating a design produces a different result for different random seeds due to the nature of \ac{rl}).
These properties make the optimization problem challenging to solve, thus requiring careful implementation.

The optimization process has two steps.
First we use \ac{bo}~\cite{frazierTutorialBayesianOptimization2018} (in practice using the package Ax~\cite{bakshyAEDomainagnosticPlatform2018}) to carry out global optimization.
\ac{bo} serves as the global exploration algorithm, efficiently sampling the parameter space to identify promising regions, and is suitable for noisy low-budget optimization.
Once \ac{bo} converges, \ac{cmaes}~\cite{hansenCMAEvolutionStrategy2016} is used to perform local refinement around the best solution found, with the purpose being to fine-tune the airframe design parameters and achieve higher performance in the final optimized solution.
As a stochastic, derivative-free optimization algorithm, \ac{cmaes} is suitable for noisy problems and iteratively improves a solution by sampling nearby solutions.

In order to speed up the optimization process, we parallelize the optimization on a high-performance cluster. 
Designs unable to produce an acceleration of $(0,0,g)$ with no angular acceleration are immediately discarded, where we set their task performance to 0 without further evaluation.

As stopping criteria, we set a limit of 750 evaluations for the \ac{bo} phase, and 250 evaluations for the CMA-ES local refinement phase. The optimization is terminated early when no improvement is observed over many consecutive evaluations. For \texttt{task A}, the optimization was stopped after \evalsBoTaskA evaluations for \ac{bo} and \evalsCMATaskA evaluations for \ac{cmaes}, totaling \totalevalsTaskA evaluations. For \texttt{task B}, the process was stopped at \evalsBoTaskB evaluations for \ac{bo} and \evalsCMATaskB evaluations for \ac{cmaes}, totaling \totalevalsTaskB evaluations.

\begin{table}[t]
	\centering
	\setlength{\tabcolsep}{4pt} 
	\caption{Motor poses for baseline designs. Arm length $r$ is scaled to match the inertia of the optimized design in each task.}
	\begin{tabular}{c|cc|cc|cc|cc|cc}
		\toprule
		motor & \multicolumn{2}{c|}{position} & \multicolumn{2}{c|}{planar} & \multicolumn{2}{c|}{franchi} & \multicolumn{2}{c|}{shen} & \multicolumn{2}{c}{rajappa} \\
		& $\phi$ & $\theta$ & $\alpha$ & $\gamma$ & $\alpha$ & $\gamma$ & $\alpha$ & $\gamma$ & $\alpha$ & $\gamma$ \\
		\midrule
		1 & 30 & 0 & 0 & 180 & 42 & 240 & 45 & 240 & 17 & 67 \\
		2 & 90 & 0 & 0 & 180 & 42 & 60 & 45 & 180 & 17 & 307 \\
		3 & 150 & 0 & 0 & 180 & 42 & 0 & 45 & 120 & 17 & 187 \\
		4 & 210 & 0 & 0 & 180 & 42 & 180 & 45 & 60 & 17 & 353 \\
		5 & 270 & 0 & 0 & 180 & 42 & 120 & 45 & 0 & 17 & 233 \\
		6 & 330 & 0 & 0 & 180 & 42 & 300 & 45 & 300 & 17 & 113 \\
		\bottomrule
	\end{tabular}
	\vspace{-1.0ex}
	\label{tab:motor_poses_baselines}
\end{table}

\subsection{Baselines}

\begin{figure}[t]
    \centering
	\begin{subfigure}{0.11\textwidth}
		\centering
		\includegraphics[width=\linewidth]{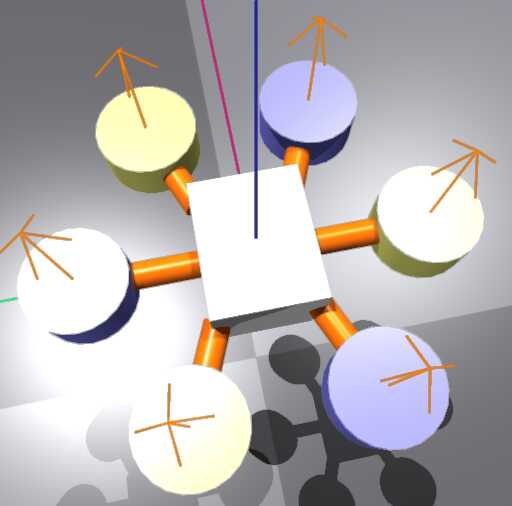}
		\caption{\emph{planar}}
		\label{fig:planar}
	\end{subfigure}
	\begin{subfigure}{0.11\textwidth}
		\centering
		\includegraphics[width=\linewidth]{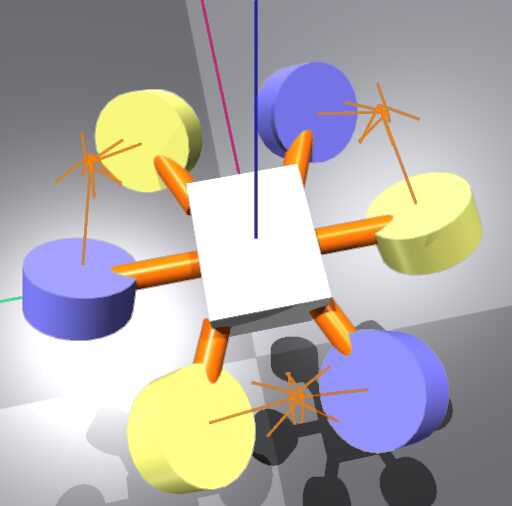}
		\caption{\emph{franchi}}
		\label{fig:franchi}
	\end{subfigure}
	\begin{subfigure}{0.11\textwidth}
	\centering
	\includegraphics[width=\linewidth]{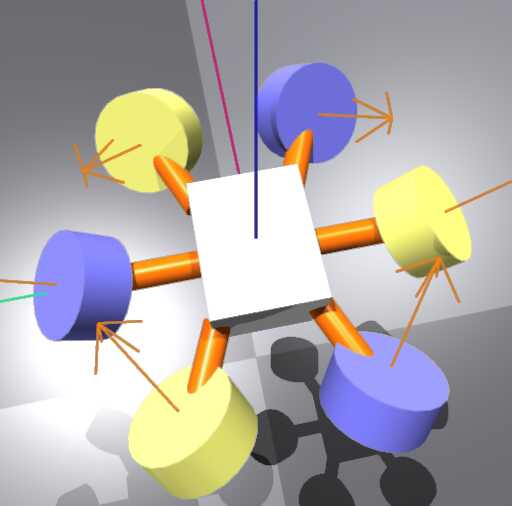}
	\caption{\emph{shen}}
	\label{fig:shen}
	\end{subfigure}	
	\begin{subfigure}{0.11\textwidth}
		\centering
		\includegraphics[width=\linewidth]{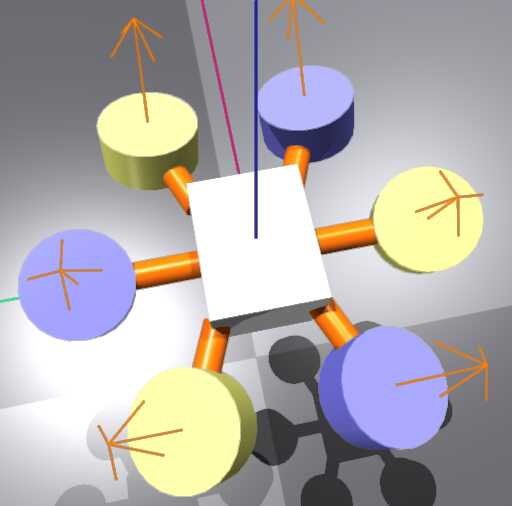}
		\caption{\emph{rajappa}}
		\label{fig:rajappa}
	\end{subfigure}
	\caption{Renders of baseline designs.}
	\label{fig:airframes}
    \vspace{-2.5ex}
\end{figure}

We include four human-engineered designs as baseline comparisons in this study.
The first design is the planar hexarotor, as shown in Figure~\ref{fig:planar}, which we denote as \emph{planar} for short in this paper.
The second and third are hexarotor designs inspired by Franchi et al.~\cite{franchiFullposeTrackingControl2018} and Shen et al.~\cite{shenModelingControlOmnidirectional2022} denoted as \emph{franchi} and \emph{shen} in this paper. They represent designs with full actuation capabilities. 
Their renders are shown in Figures~\ref{fig:franchi} and \ref{fig:shen} respectively.
Finally, we include the design inspired by Rajappa et al.~\cite{rajappaModelingControlDesign2015} (\emph{rajappa}) also with multidirectional actuation, but the tilting angle is less aggressive than in the previous two (Figure~\ref{fig:rajappa}).
The poses of the motors of the baseline designs are shown in Table~\ref{tab:motor_poses_baselines}. 
All baseline designs are closely matched to the optimized designs in terms of mass and inertia.
We achieve this by setting the arm length of the baseline designs to the average arm length of the optimized design in each task.

All designs (baseline and optimized) undergo the exact same training and evaluation process: identical \ac{rl} procedure, reward functions, and task performance measurement.
To try to get the most performance out of each baseline and optimized design, we increase the number of starting policies in the sequential halving procedure described in Section~\ref{sec:sequential_halving} to 32.

%

\section{Evaluation Studies in Simulation}
\label{sec:experimentation}

\begin{figure*}

	\centering
	\includegraphics[width=0.87\linewidth]{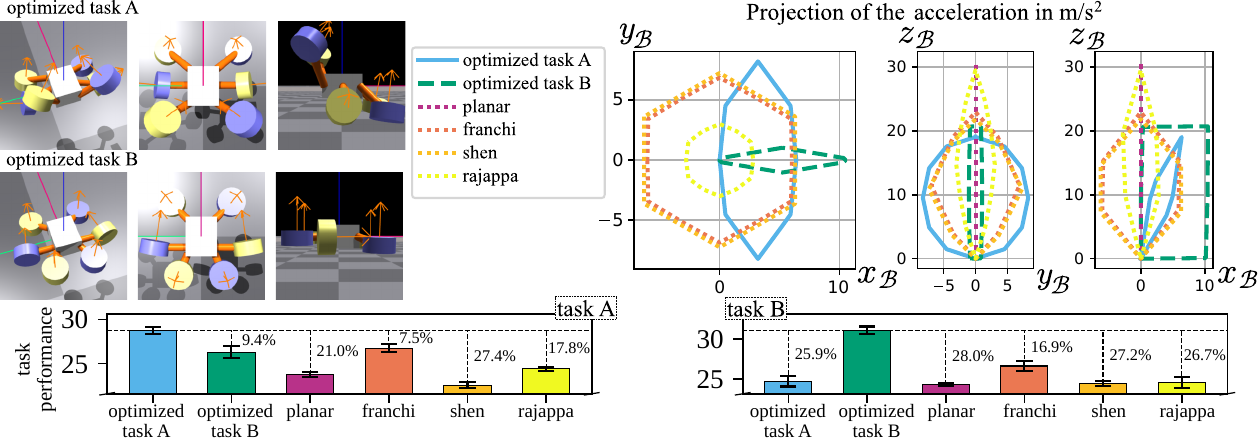}
	\caption{\emph{Top left}: Different views of the optimized designs. Orange arrows show the direction of the force generated by the motor.  \emph{Top right}: projection of the limits of linear acceleration for the optimized design and the baselines. \emph{Bottom}: The task performance of the optimized design and the baselines for 8 different training runs each. Percentage represents performance loss with respect to the best performing design in that task (lower is better).}
	\label{fig:results_renders_simulation_both_tasks}
    \vspace{-4ex}
\end{figure*}

\subsection{Design optimized for \texttt{task A}}
The renders of the optimized design for \texttt{task~A} are shown in the top left side of Figure~\ref{fig:results_renders_simulation_both_tasks}.
This design has four propellers tilted forward, with two other propellers facing towards the sides.
When compared with the baseline designs, this design can accelerate faster in $y$ as can be seen in the maximal acceleration projections in Figure~\ref{fig:results_renders_simulation_both_tasks}. 

To compare the task performance of the airframes, we train them 8 times each and show the average and standard deviation in the bottom part of Figure~\ref{fig:results_renders_simulation_both_tasks}.
The optimized design obtains a \percentageTaskAOptimizedBetterThanPlanar higher task performance than the planar design, and \percentageTaskAOptimizedBetterThanFranchi higher than the best out of the other baseline designs.
See \href{https://youtu.be/V6w_DTKWvtc?t=88}{the video associated to the paper at minute \videoexptimestampone} for a comparison of the optimized design with the baselines in \texttt{task A} in simulation.

\subsection{Design optimized for \texttt{task B}}

As shown in the top left part of Figure~\ref{fig:results_renders_simulation_both_tasks}, this optimized design has four propellers pointing upward and two propellers pointing forward. When flying forward, the \ac{mav} tilts backward, where four of the motors can produce force backward, and two motors are producing force forward and countering gravity. This arrangement of the motors is also evident in the $z_{\mathcal{B}}-x_{\mathcal{B}}$ projection of the maximal acceleration in Figure~\ref{fig:results_renders_simulation_both_tasks}. This gives the drone a shorter stopping distance, hence allowing for a faster nominal speed when flying forward on \texttt{task B}, as compared to the baseline \ac{mav} designs. This is important for \texttt{task B}, where a stopping distance of less than $0.25$~m is required to clear waypoints without missing any when there is a change in $y$, and the task otherwise involves mostly moving forward. 
The optimized design achieves a task performance \percentageTaskBOptimizedBetterThanPlanar higher than the planar design, and \percentageTaskBOptimizedBetterThanFranchi higher than the best out of the other baseline designs (bottom right part of Figure~\ref{fig:results_renders_simulation_both_tasks}).
See also \href{https://youtu.be/V6w_DTKWvtc?t=131}{the paper video at minute \videoexptimestamptwo} for a comparison of the optimized design with the baselines in \texttt{task B} in simulation. 

\begin{figure}
	\centering
	\includegraphics[width=0.76\linewidth]{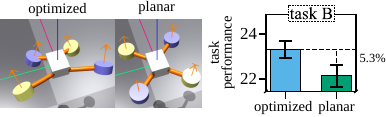}
	\caption{\textit{Left:} Renders of the optimized and planar quad. \textit{Right:} Task performance of the quadrotors in \texttt{Task B}.}
	\label{fig:quadfigure}
\vspace{-2.2ex}
\end{figure}

\emph{Optimizing quadrotors:} we also tested the methodology in the quadrotor case in \texttt{task B}. 
The experimental setup for the quadrotor was exactly the same as the hexarotor except in the number of motors.
The optimized quadrotor obtains a task performance \percentageTaskBOptimizedBetterThanPlanarQuad higher than a standard quadrotor when matched for mass and inertia (see Figure~\ref{fig:quadfigure}).
However, the difference with respect to the baseline was smaller in this case. We hypothesize that this could be due to the reduced set of flying configurations available in the quadrotor case. 

\emph{Cross task performance:}
We evaluate the \textit{optimized task A} design on \texttt{task B} and vice versa (bottom right part of Figure~\ref{fig:results_renders_simulation_both_tasks}).
The \textit{optimized task B} design performs relatively well compared to the baselines in \texttt{task A}, only 2\% worse than the best out of the baselines (the \textit{Franchi} design).
The \textit{optimized task A} design also outperforms all baseline designs except \textit{Franchi} in \texttt{task B}, but this time the performance is closer to the poor performing baselines.
We conclude from this experiment that the performance of the optimized designs is on par with the rest of the non-specialized designs for the forward flying task that they were not optimized for.

\subsection{Global optimization and local refinement}

\begin{figure}
	\centering
	\includegraphics[width=1.0\linewidth]{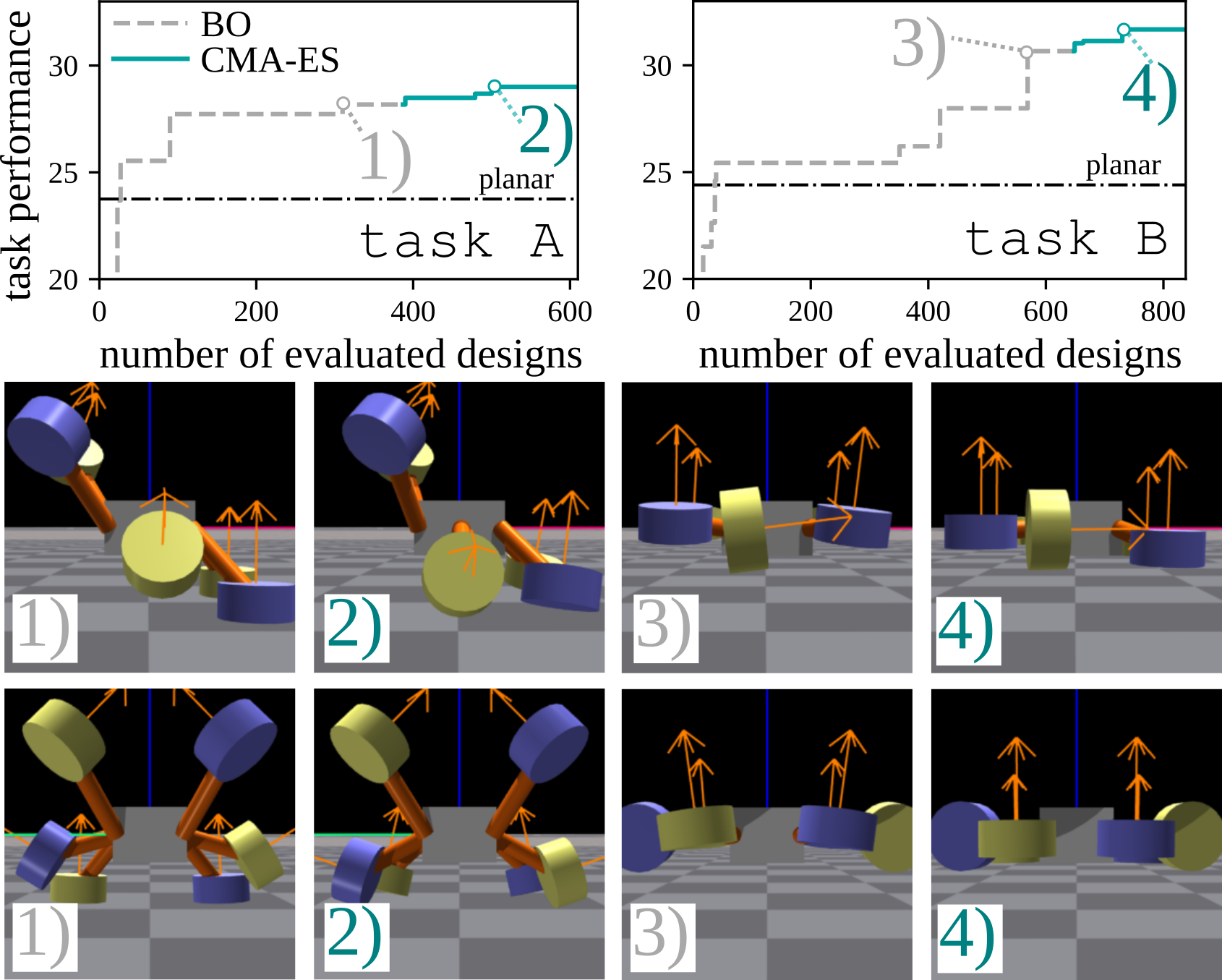}
	\caption{Task performance of best found design with respect to the number of evaluated designs, with the performance of planar design added as reference (top). Renders show design refinements: {\color{mygrey}1), \color{myteal}2)} \texttt{task A} designs before and after CMA-ES showing more aggressive motor tilt angles; {\color{mygrey}3), \color{myteal}4)} \texttt{task B} designs before and after CMA-ES showing improved propeller alignment with coordinate axes (bottom).}
	\label{fig:sobolbocmaestaskab}
    \vspace{-2.5ex}
\end{figure}

As mentioned before, the optimization procedure considered in this paper starts with \ac{bo} followed by \ac{cmaes} that starts with the best found solution of \ac{bo}.
The top part of Figure~\ref{fig:sobolbocmaestaskab} shows the task performance as the number of evaluated solutions increases, with the performance of the planar design as a reference.
The largest contribution in task performance comes from \ac{bo}, while the local refinement by CMA-ES provides an incremental improvement.
We highlight the best designs found by \ac{bo} only and \ac{bo}~+~\ac{cmaes} in the bottom part of Figure~\ref{fig:sobolbocmaestaskab}.
In \texttt{task A} the local refinement makes the tilt angle of the center and front motors more aggressive, as can be seen by comparing the renders {\color{mygrey}1)} and {\color{myteal}2)} in the bottom part of Figure~\ref{fig:sobolbocmaestaskab}, with an improvement of $\percentageCmaesAfterBoTaskA$ in task performance.
In \texttt{task B}, CMA-ES flattens out the angles of the propellers to be better aligned with the $z$ and $x$-axes, as can be seen by the difference in renders {\color{mygrey}3)} and {\color{myteal}4)}. This improves task performance by $\percentageCmaesAfterBoTaskB$.

\subsection{Sensitivity to Task Variation}

\begin{figure}
\vspace{1.5ex}
    \centering
    \includegraphics[width=1.0\linewidth]{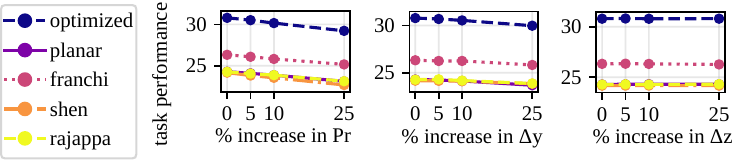}
    \caption{Median task performance for \texttt{task B} task variations. Note the lines of \emph{planar}, \emph{shen} and \emph{rajappa} largely overlap.}
    \label{fig:corrupted_tasks_results}
\vspace{-1.5ex}
\end{figure}

The previous results demonstrate that optimized designs outperform the baselines for the tasks for which they were optimized. 
We now proceed to assess robustness against task variation. 
Specifically, we increase the difficulty of \texttt{task~B} (task related to last-resort emergency obstacle avoidance) by increasing the parameters $\mathrm{Pr}$, $\Delta y$, and $\Delta z$ as defined in Section~\ref{sec:task_definition} by up to 25\%.
Increasing $\Delta y$ or $\Delta z$ makes the task more difficult because it increases the cross-track distance from the previous waypoint to the next when a maneuver is required, and so does increasing the probability $\mathrm{Pr}$. 
We evaluate the airframes again (with the same policies) and show the median task performance in Figure~\ref{fig:corrupted_tasks_results}.

As expected, there is a drop in performance for all designs when increasing $\mathrm{Pr}$ and $\Delta y$. There is no drop in performance when changing $\Delta z$ because the size of the maneuver that the drone needs to do in $z$ remains relatively small.
Importantly, the optimized design follows a similar trend to the baselines where the performance of all of the designs decreases by a similar amount.
This shows that airframe optimization and learned policies remain robust to task variations and the optimized design maintains its advantage over baselines across these variations.

\section{Sim2Real Transfer}
\label{sec:sim2Real}

We built and tested the optimized drone for \texttt{task B} to see how well the simulation results transfer to the real world.
We chose \texttt{task B} for sim2real experimentation because both the distribution of the waypoints and the optimized airframe are more unconventional than the baseline designs. 
We went for a 3D-printed modular design with interchangeable arms.
The modular design allows quick swapping between different designs without having to change the electronics.
We use the XING 1404 3000KV motors with HQProp Duct-T75MMX5 75mm propellers.
We test the airframe in a $6 \times 6$~m arena, where the state of the robot is estimated with a Qualisys Motion Capture system and the IMU.
We run the neural policy onboard at 200~Hz using ModalAI VOXL2 Mini. 
Policy inference is integrated with the PX4 autopilot firmware.


%

As a refinement step, the policy deployed in the real world is slightly modified compared to the policy used during the simulation-based optimization process. 
First, given that we have the real drone and the CAD model, we can estimate the mass and inertia more accurately compared to the way this is calculated in Aerial Gym.
We update these values before retraining the policy.
Moreover, to ensure safer flight, we also introduce the smoothing loss of Mysore et al.~\cite{mysoreRegularizingActionPolicies2021} to get smooth motor commands.
This reduces angular acceleration and unnecessary oscillations.

As expected, there is a trade-off between the smoothness in policy actions and the task performance of the airframe. 
Enabling the smoothing penalty reduces the task performance of the produced policies and the associated airframe designs.
The task performance of the optimized design drops by $10\%$ from $30.9$ to $27.8$, which is still higher than the best of the baselines even without action smoothing (see Table~\ref{tab:smoothing_vs_no_smoothing}).
Baseline designs also show a $5-10\%$ performance drop.
Still, the results of our airframe-policy optimization process hold: the optimized design keeps outperforming the baselines both when those use or do not use action smoothing.

\begin{table}[h!]
\centering
\vspace{-0.5ex}
\caption{Effect of action smoothing on task performance in \texttt{task B}.}
\label{tab:smoothing_vs_no_smoothing}
\begin{tabular}{lccc}
\hline
\textbf{Design} & \textbf{No smoothing} & \textbf{Smoothing} & \textbf{Performance Drop (\%)} \\
\hline
optimized & \textbf{30.9} & \textbf{27.8} & 9.9 \\
planar    & 24.3 & 22.2 & 8.6 \\
franchi   & 26.6 & 24.2 & 9.1 \\
shen      & 24.4 & 23.0 & 5.7 \\
rajappa   & 24.5 & 22.1 & 9.8 \\
\hline
\end{tabular}
\vspace{-1ex}
\end{table}

Finally, before real robot deployment, we use a second-order motor model (with rate limiter) that better tracks the real actuation capabilities of the XING 1404 3000KV motors that we use with the HQProp Duct-T75MMX5 75mm propellers.
To fit the second-order motor model parameters, we first deploy the real drone with the policy trained in simulation with the simpler first-order motor model, and then we fit the parameters of the second-order motor model such that the difference in \ac{rps} between reality and simulation is minimized.
In the following, we show the results of deploying the airframe optimized for \texttt{task B} in the real world.

\subsection*{Deploying the airframe optimized for task B in the real world}

\begin{figure}
    \vspace{1ex}
	\centering
	\includegraphics[width=1.0\linewidth]{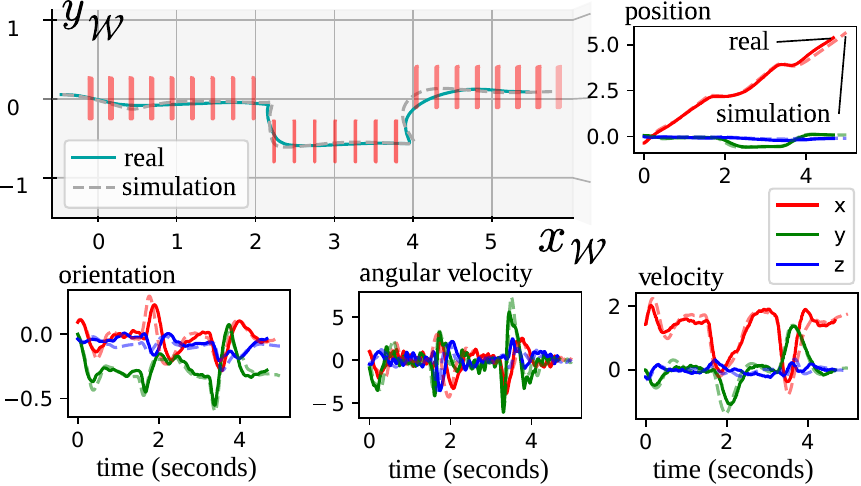}
    \caption{\texttt{task B} real and simulated trajectories.} 
	\label{fig:realvssimtopview_task_B}
    \vspace{-2ex}
\end{figure}

The real and simulated trajectories are shown in Figure~\ref{fig:realvssimtopview_task_B}.
The real world experiment is shown in the \href{https://youtu.be/V6w_DTKWvtc?t=176}{paper video at minute \videoexptimestampthree}.
As shown in the figure, the real drone follows a very similar trajectory to the simulated one.
Looking at the positions over time (left-top), we can see that the performance in reality matches the simulation.
The rest of the values, like the velocity, also match well with the simulation.
This experiment shows that optimized designs, even when rather unconventional, are feasible in the real world, and that the control behavior is comparable to that of simulation.

\section{Conclusions}
\label{sec:conclusion}

In this paper, we optimize hexarotor \ac{mav} designs for particular navigation tasks, defined as waypoint distributions, guided only by task performance.
With \ac{bo}, \ac{cmaes}, and \ac{rl}, we introduce a framework to iteratively improve on the task by sampling new airframe designs, training them, and testing how well they perform.
Simulated studies show significant improvements both against planar hexarotors and tilted designs from the literature (matched for mass and inertia).
Real-world experiments show that the airframe behavior transfers well to reality. Overall, this work demonstrates the value of task-specific robot design and control co-optimization guided by closed-loop performance evaluation, validated by real-world experiments. Future work shall expand to include (a) co-optimization of motor poses and motor-propeller combinations, (b) elastic elements in the airframe (e.g., as in human-engineered designs~\cite{mintchev2017insect}), as well as (c) deciding sensor placement.

%





\bibliographystyle{IEEEtran}
\bibliography{main}

\end{document}